\documentclass{article}

\usepackage[dblblindworkshop, final]{neurips_2026}
\workshoptitle{Neural Network Artifacts as a New Data Modality}

\usepackage[utf8]{inputenc} 
\usepackage[T1]{fontenc}    
\usepackage{hyperref}       
\usepackage{cleveref}
\usepackage{url}            
\usepackage{booktabs}       
\usepackage{multirow}
\usepackage{amsfonts}       
\usepackage{nicefrac}       
\usepackage{microtype}      
\usepackage{xcolor}         
\usepackage{graphicx}
\usepackage{float}  

\title{Localizing Transfer Between Memorization Tasks}

\author{%
  Yimiao Yu \\
  Center for Data Science \\
  New York University \\
  \texttt{yy5915@nyu.edu} \\
  \And
  Florentin Guth\thanks{Work done while partially affiliated with NYU CDS.} \\
  Center for Computational Neuroscience \\
  Flatiron Institute \\
  \texttt{fguth@flatironinstitute.org} \\
}

\begin{document}

\maketitle

\begin{abstract}
   A central puzzle in transfer learning is why pre-training on one task can accelerate training or improve performance on another task, and what mechanisms underlie this transfer. In this work, we examine the transfer between memorization tasks of random input-output mappings. We find two surprising transfer patterns: \emph{equivalent transfer}, where each additional pre-training epoch saves approximately one downstream fine-tuning epoch; and \emph{non-equivalent transfer}, where pre-training on a mismatched task can be even more efficient than directly training on the downstream task itself. Through ablation experiments, we decompose and localize the transfer into two separate effects: a ``trivial'' magnitude-driven transfer in the last layer, and a ``non-trivial'' structure-driven transfer, partially attributable to the \emph{covariance} of the other layers. These results advance our understanding of the underlying mechanisms of transfer learning and have the potential to lead to principled pre-training strategies.
\end{abstract}

\section{Introduction}

Standard network training starts from an uninformed, random initialization of the network parameters. In contrast, transfer learning initializes the weights from a learned checkpoint obtained from an \emph{upstream} pre-training task, before fine-tuning them on the \emph{downstream} task of interest. This setup may be further modified in a variety of ways, by freezing or resetting some layers, reducing the learning rate, etc. \citep{yosinski2014transferable,donahue2014decaf,howard2018ulmfit,kornblith2019better,zhang2022layers,li2018explicit}. Transfer learning is often used when there is little downstream data, as it can make use of a potentially much larger pre-training dataset. The effectiveness of the transfer can be measured in two main ways, via either final test performance or time to convergence compared to a random initialization. A successful transfer presumably requires the pre-training and downstream tasks to be sufficiently similar so that solving them requires learning a common set of ``features''. But can this be made more precise? How should the effectiveness of transfer be evaluated, when does it occur, what do the two tasks need to share? Can we provide a mechanistic explanation for transfer, locate where and how the learned ``features'' are encoded in the weights, and which part should be kept or removed during transfer? Answering these questions will inform the design of the pre-training task and transfer mechanism to maximize downstream performance and training speed.

In order to make progress on these questions, we first note that, in assessing transfer, one needs to disentangle \emph{optimization} (the training dynamics) from \emph{generalization} (as measured by test loss or another performance metric of interest). We will limit ourselves to the former here, and study memorization rather than generalization. This enables us to generate many synthetic tasks by randomizing labels or image pixels (inspired by \citet{zhang2017understanding}, who showed that standard architectures fit random labels and random pixels as readily as real data), and thereby change independently both the target function to learn $y = f(x)$ and the input distribution $p(x)$.

Our results reveal a rich picture with two quantitative transfer effects. Prior work has shown pre-training on random labels can accelerate the downstream fine-tuning \citep{pondenkandath2018leveraging, maennel2020what} but did not quantify the transfer effect. We show quantitatively that there is a surprising one-to-one equivalence between the pre-training and fine-tuning epochs when upstream and downstream tasks share the same image distribution, which we call ``\emph{equivalent transfer}''. Transfer across different distributions is itself common practice (e.g., from ImageNet to medical images \citealp{shin2016deep,ke2021chextransfer}), and is typically attributed to ``reusable features'' \citep{yosinski2014transferable} or ``low level statistics of data'' \citep{neyshabur2020transferred}. We show that these are not necessary: we further identify ``\emph{non-equivalent transfer}'', where pre-training on a task devoid of structure (random pixels, whose pixel values differ from natural images) can be even more efficient than directly training on the downstream task. This challenges the covariance alignment explanation by \citet{maennel2020what}, where they implicitly assume the data distributions of upstream and downstream are identical. Beyond identifying these transfer patterns, we further investigate whether these effects can be simulated more ``cheaply'', and find that the upstream label consistency (fixed labels) is essential for the transfer, while data consistency (fixed pre-training images) is not, suggesting a path towards simpler and more efficient pre-training strategies.

We also seek to localize the transfer within the network weights. Through ablation experiments, we decompose the transfer effect into two distinct parts: magnitude-driven transfer and structure-driven transfer. The magnitude-driven transfer is fully attributed to the growing weight scale, while we partially attribute the structure-driven transfer to the weight covariance. We publish code at \url{https://github.com/feima3333/transfer-between-memorization-tasks}.

\section{Equivalent and non-equivalent transfer in memorization tasks}
\label{transfer phenomenon}

\begin{figure}[t]
  \centering
  \includegraphics[width=0.4\linewidth]{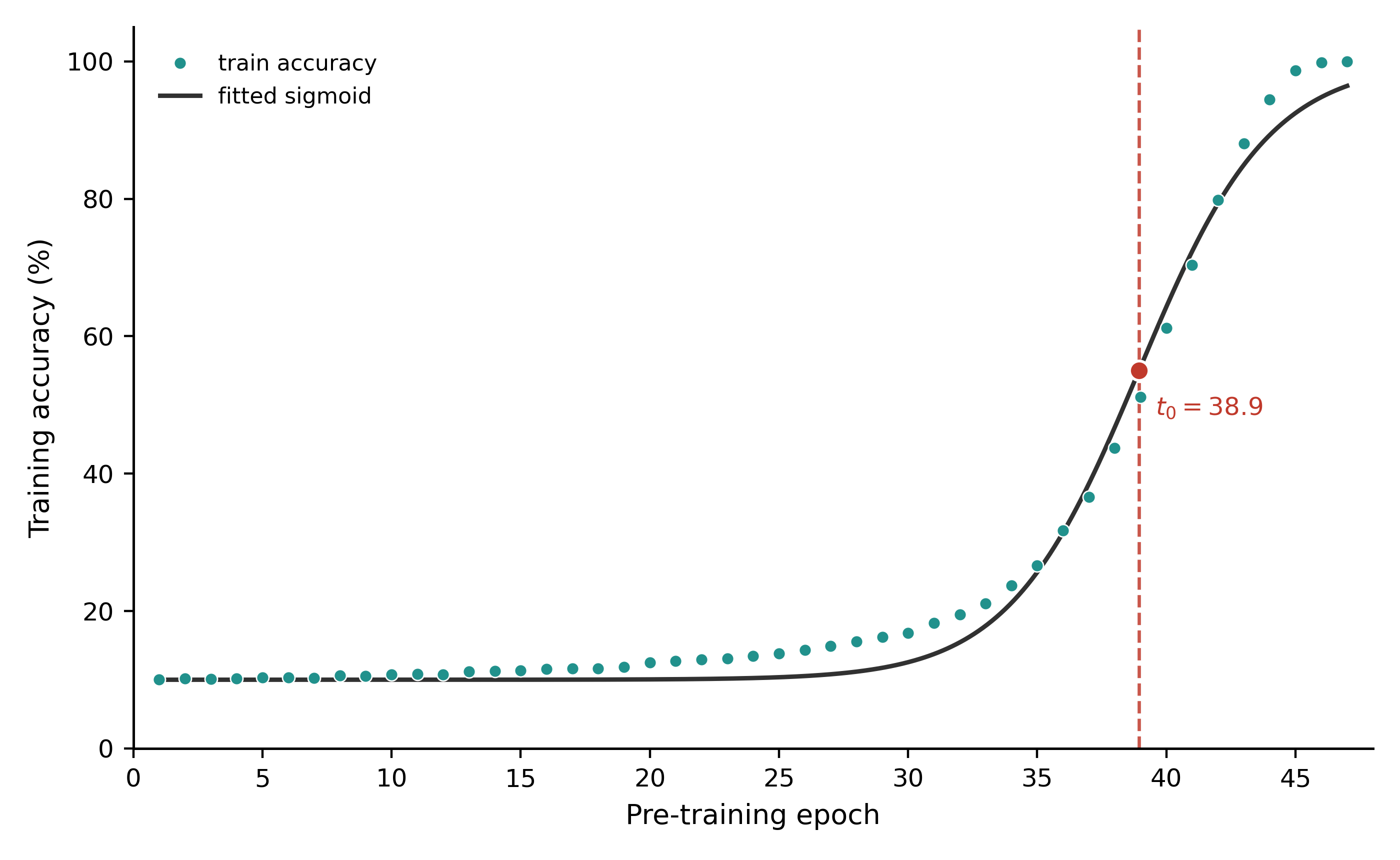}
  \caption{The training curve of a random label task is well-approximated by a sigmoid and can be summarized by its inflection point $t_0$. We plot training accuracy (green dots) as a function of training time, and its fitted sigmoid (black line) defines the effective training time $t_0$ (dashed red vertical line) at its inflection point.} 
  \label{fig:t0_definition}
\end{figure}

We initiate a scientific study of transfer learning: how do we quantify it, when does it occur, what do the upstream and downstream tasks need to share? To do so, we introduce a controlled experimental setting where we investigate the effect of different pre-training tasks on random label fine-tuning (each image is assigned a random label independently of its content). For each pre-training task, we evaluate fine-tuning time to reach convergence as a function of pre-training time, by generating a sequence of checkpoints along the pre-training trajectory and using each one as initialization for the downstream task. We empirically observe that when training on random labels, the training accuracy curve as a function of training time is well-approximated by a sigmoid, as pictured in \Cref{fig:t0_definition}. Since training accuracy always starts at chance level and eventually reaches 100\%---the networks we consider have enough capacity to fit the training set---the entire curve can be summarized by its inflection point $t_0$, which defines an effective training time, and its slope. We will focus on $t_0$ for simplicity, and define transfer as occurring when pre-training yields a reduced $t_0$ compared to training from scratch. Throughout this work, we use a standard CNN with six convolutional layers, where each convolutional layer is followed by batch normalization and ReLU, and a fully connected classifier. We include more details on the experimental setting in Appendix \ref{app:exp_details}.

\begin{figure}[t]
  \centering
  \includegraphics[width=\linewidth]{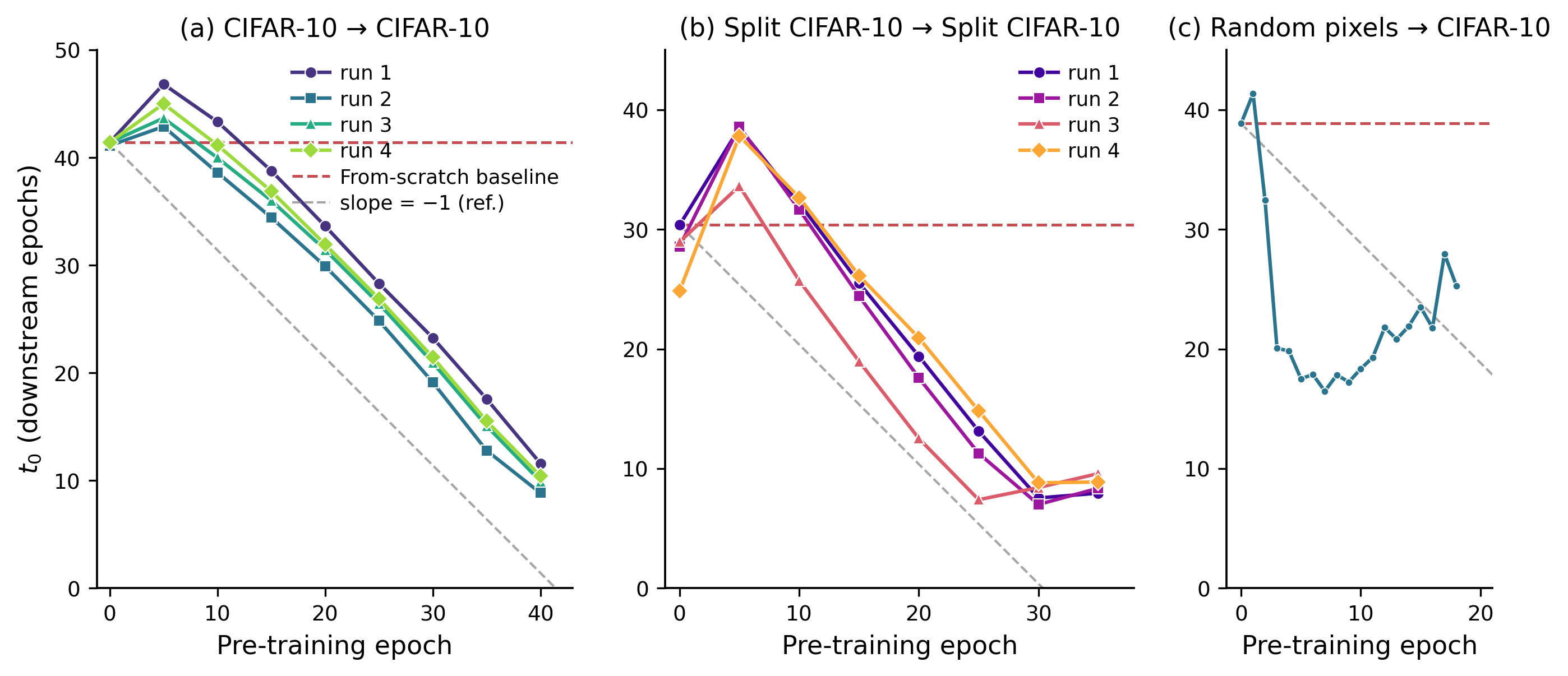}
  \caption{Equivalent and non-equivalent transfer from different pre-training tasks to random label classification. Each panel shows fine-tuning time (as measured by $t_0$) as a function of pre-training time. \textbf{(a)} Transfer between identical images. All four curves, corresponding to different pre-training random label mappings, are almost parallel to the gray reference line (slope = $-1$) after a warm-up period of $\approx 5$ epochs: we call this phenomenon \emph{equivalent transfer}. \textbf{(b)} Transfer between shared image distribution. We split the training set with randomized labels into two halves, and pre-train on the first and fine-tune on the second. We find that equivalent transfer roughly persists across four different splits, although $t_0$ decreases by slightly more than $1$ with each additional pre-training epoch.
  \textbf{(c)} Transfer across severely mismatched distribution. We pre-train on random pixel images. We find that a different, \emph{non-equivalent transfer} is taking place: for a range of epochs, $t_0$ after random pixel pre-training is smaller than the reference $-1$ slope line. There is a net acceleration: pre-training is here more efficient than fine-tuning.
  } 
  \label{fig:t0_linear}
\end{figure}

We consider three pre-training tasks with varying amounts of shared structure between the upstream and downstream tasks. We show in \Cref{fig:t0_linear} that quantitative transfer persists even as the train and test tasks become more different. In panel (a), when the upstream and downstream tasks share identical training images, we find that $t_0$ decreases with pre-training epochs at a near-unit slope (i.e., each additional pre-training epoch saves us one fine-tuning epoch), after a warm-up period of $\approx 5$ epochs. This phenomenon, which we call \emph{equivalent transfer}, is robust across four random label seeds. This is surprising: not only is it not clear why memorizing one set of random labels should help with memorizing another set of random labels, but even less at a rate where one upstream epoch is equivalent to one downstream epoch (after the warm-up period). In panel (b), we further demonstrate that the equivalent transfer persists even when pre-training and fine-tuning occur on different images (each with random labels) by splitting the training set into two disjoint sets. In panel (c), to test whether the transfer effect holds across severely mismatched image distributions between upstream and downstream tasks, we pre-train on a random pixel dataset: each pixel has an i.i.d.\ uniform color, with the same size as CIFAR-10 training set. We find a form of ``\emph{non-equivalent transfer}'', with $t_0$ dropping significantly within the first few epochs, followed by a plateau and a slight increase. Remarkably, there is a range of pre-training epochs for which it is \emph{more efficient} to train on random pixels than the downstream task, although there is in a sense ``nothing to learn'' from random pixels.

\begin{figure}[t]
  \centering
  \includegraphics[width=\linewidth]{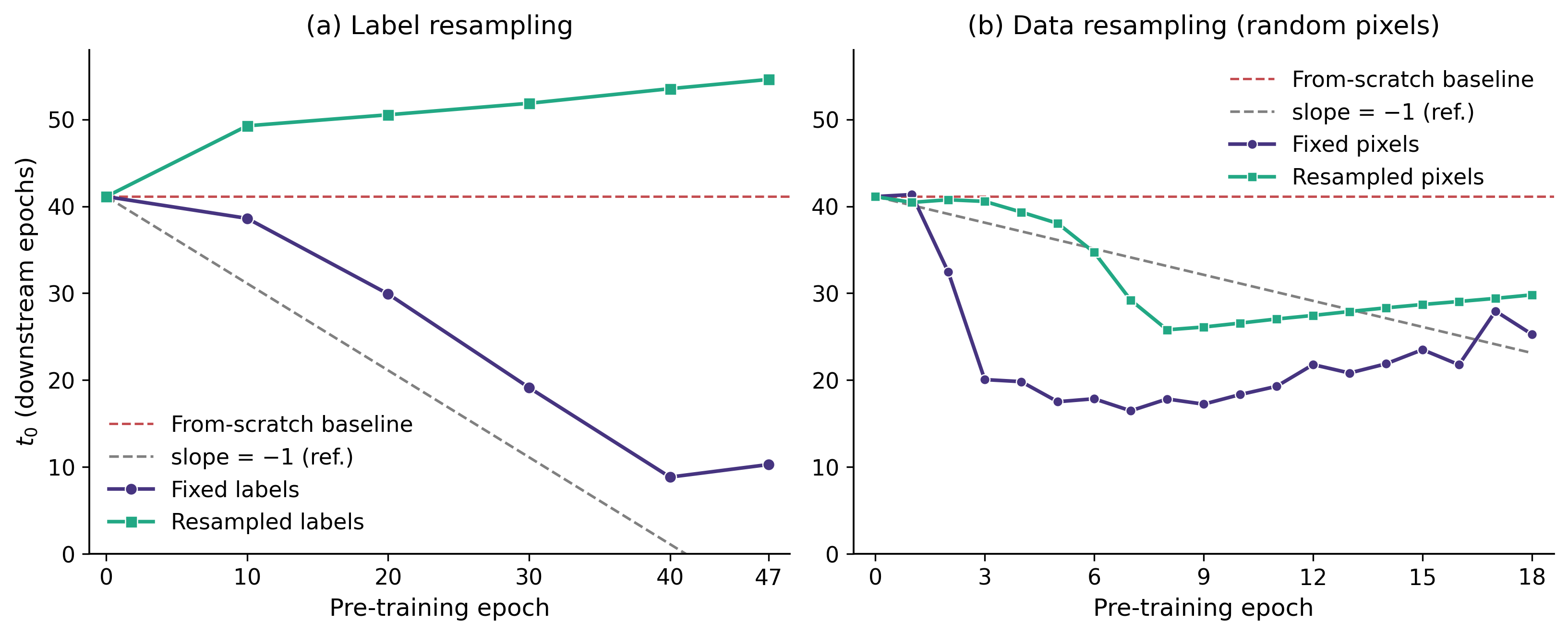}
  \caption{Effects of resampling labels or input images on transfer to memorization of random labels on CIFAR-10 images. 
  \textbf{(a)} Label resampling: during pre-training, random labels are resampled at the end of each epoch (teal) compared with a fixed set as in \Cref{fig:t0_linear} (a) (purple). Under this resampling setting, $t_0$ is consistently larger than the from-scratch baseline.
  \textbf{(b)} Data resampling: during pre-training, random pixel images are resampled at the end of each epoch (teal) compared with a fixed set as in \Cref{fig:t0_linear} (c) (purple). Non-equivalent transfer is weakened but persists. }
  \label{fig:resampling comparison}
\end{figure}

The fact that transfer can occur between synthetic tasks that seemingly share no common features raises the question of whether one even needs a consistent (fixed) pre-training task. Can we simulate the effects of pre-training in a simpler way? We now test if label consistency and data consistency across epochs are necessary by conducting two resampling experiments in \Cref{fig:resampling comparison}. In panel (a), we resample the random labels after each pre-training epoch. This removes label consistency by changing the target function at each epoch. We find that it prevents any kind of transfer, equivalent or not, although each random label seed in isolation leads to equivalent transfer (compare with panel (a) of \Cref{fig:t0_linear}). In contrast, in panel (b), we observe that non-equivalent transfer persists under data resampling, where random pixel images are resampled after each pre-training epoch. This is surprising as each training example is seen only once: each batch is a fresh sample of random data and pre-training accuracy remains at chance level indefinitely. But unlike label resampling, where successive epochs conflict (the same image is associated with different labels), the data resampling experiment can be considered as seeing different realizations of the same task, and thus involves no conflict. This raises the question whether a similar experiment would preserve equivalent transfer between random label tasks without repetition, but this would require access to ``independent identically-distributed copies'' of CIFAR-10, e.g., sampled from a class-conditioned generative model.

Our results show that transfer can occur without shared data between the upstream and downstream task, and even when the pre-training loss curve is flat (as in data-resampling memorization tasks). However, the specific form and efficiency of transfer depends on the data distribution, with equivalent transfer seemingly requiring a shared distribution.

\section{Locating the source of transfer: ablation experiments}
\label{sec:where}

\begin{figure}[t]
  \centering
  \includegraphics[width=\linewidth]{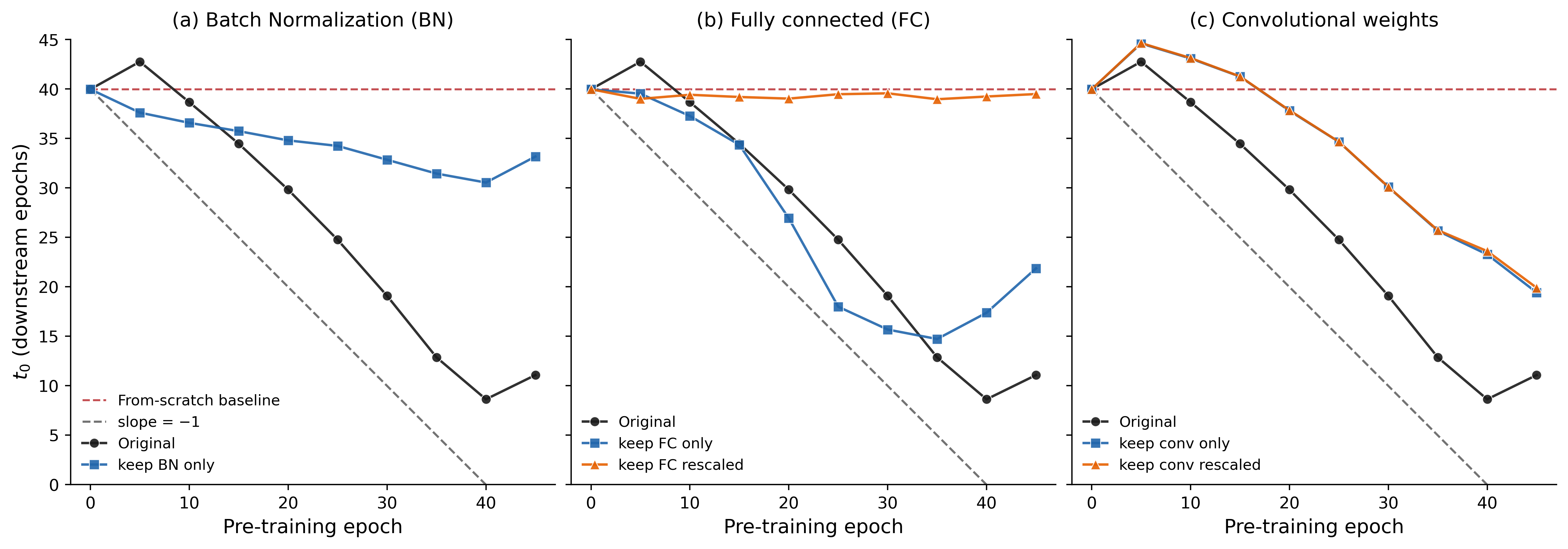}
  \caption{Transfer between random label tasks while resetting all but one type of weights (blue), keeping only either batch normalization affine parameters \textbf{(a)}, fully-connected weights \textbf{(b)}, or convolutional weights \textbf{(c)}. We find that the transfer effect only persists when keeping FC or convolutional weights when compared to the full pre-trained checkpoint (black). We also include checkpoints where the global magnitude (norm) of the transferred weights has been reset to its initialization value by adjusting a single scale factor per layer (orange).  The transfer carried by FC weights can be fully explained by their growth in magnitude, in contrast to the convolutional weights.} 
  \label{fig:weight_ablation_randlabel}
\end{figure}

In this section, we seek a mechanistic explanation of transfer: where and how are the learned ``transferable features'' encoded in the pre-trained weights?
In our model, there are only three types of learned parameters: convolutional weights (all layers except the last one), fully-connected (FC) weights (the last layer), and the affine parameters of batch normalization layers \citep{ioffe2015batchnorm}. To locate where the transfer may lie, we introduce a weight ablation experiment: from a given pre-trained checkpoint, we reset all weights but one type to their initialization, and start fine-tuning the partially reset checkpoint on the downstream task. Results in the random label setting of \Cref{transfer phenomenon} are shown in \Cref{fig:weight_ablation_randlabel}, and results for random pixels are shown in \Cref{fig:weight_ablation_random_pixel_aligned} in Appendix \ref{sec:supplementary results}. We find in both settings that batch normalization affine parameters alone carry little transfer, while keeping either FC weights or convolutional weights preserves a substantial transfer effect. 

To understand what drives this transfer, we first compare the magnitude of each weight type between initialization and after pre-training in \Cref{tab:norm_growth}. 
Under both random labels and random pixels pre-training, batch normalization weights remain nearly at initialization scale, convolutional weights grow modestly, and FC weights grow by more than an order of magnitude. This raises the question whether transfer is simply caused by this norm growth: with a fixed learning rate, weights with larger norm generate larger gradient updates, which can accelerate downstream fine-tuning (i.e., reduce $t_0$).

To test whether the FC transfer and conv transfer are driven by the growing magnitude of the weights, we introduce a rescaling experiment: from a given pre-trained checkpoint (on a random label task), we keep only the FC/convolutional weights and also rescale their norm to its initialization value. The results are shown in panels (b) and (c) of \Cref{fig:weight_ablation_randlabel}. After rescaling, the FC transfer disappears entirely, indicating that the FC transfer can be fully explained by the ``trivial'' norm growth. This demonstrates that weight scale is an important confounder in transfer learning experiments, and should always be accounted for in future investigations.
In contrast, the rescaled checkpoint behaves nearly identically for the conv transfer experiment, showing that it cannot be explained by the growth in norm.
We also obtained similar results in the setting of random pixels pre-training (see \Cref{fig:weight_ablation_random_pixel_aligned} in Appendix \ref{sec:supplementary results}).

\begin{table}[t]\centering
\caption{Frobenius norm ratios of different weights between initialization and after training. Norms are computed over all relevant parameters, including bias parameters, and across layers if applicable.
We observe that in both random label and random pixel settings, fully-connected weights grow by more than an order of magnitude, while convolutional and batch-normalization weights remain near the initialization scale.}

\label{tab:norm_growth}
\begin{tabular}{rrr}
\toprule
\multirow{2}{*}{\textbf{Weight type}}  & \textbf{Random labels} & \textbf{Random pixels} \\
& (epoch $0 \rightarrow 45$) & (epoch $0 \rightarrow 15$) \\
\midrule
Batch normalization & $\times$1.05 & $\times$1.01 \\
Convolutional & $\times$1.14 & $\times$1.19 \\
Fully-connected & $\times$23.70 & $\times$14.40 \\
\bottomrule
\end{tabular}
\end{table}

Since the convolution transfer cannot be explained by the magnitude of the weights, we shift our attention to more fine-grained properties. \citet{maennel2020what} find that, in the random label pre-training setting, preserving the weight covariance structure can accelerate the downstream fine-tuning considerably, across several architectures. \citet{guth2024rainbow} show that a Gaussian model preserving only the weight covariance at each layer (with a suitable activation alignment) can well approximate the upstream performance of a trained network. These works suggest that the convolution transfer is encoded in the covariance matrices of convolutional layers. To test this hypothesis, we factorize weights as a product of covariance square root and whitened weights, and reset each individually.

\begin{figure}[tbp]
  \centering
  \includegraphics[width=\linewidth]{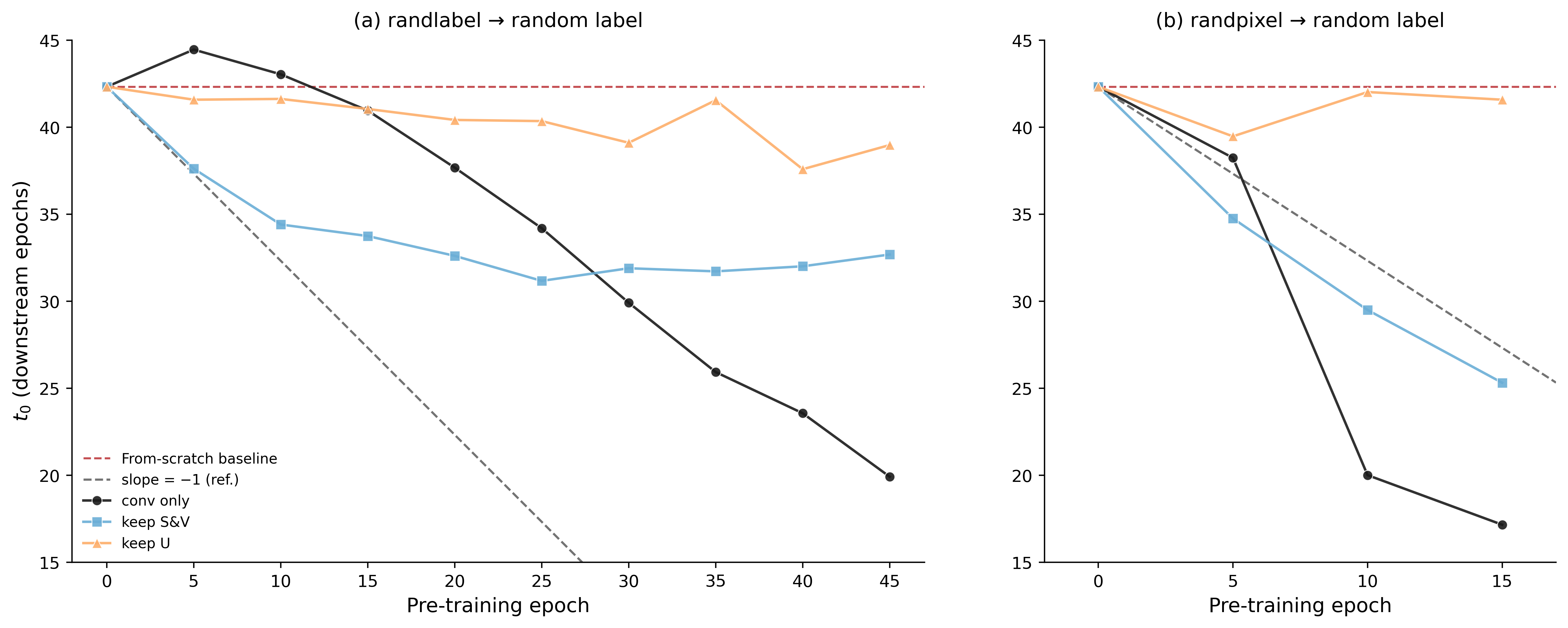}
  \caption{Transferring SVD factors of the convolutional weights (batch normalization affine parameters and FC weights are reset to their initialization). \textbf{(a)} In the random label setting, the transfer effect completely disappears after resetting $S$ and $V$ (thus resetting $C$), but some transfer remains after resetting $U$ (thus keeping $C$).  Notably, keeping $C$ reduces $t_0$ during early pre-training ($\approx 10$ epochs), then plateaus while conv-only checkpoints continue to decrease $t_0$. \textbf{(b)} We observe the same pattern for non-equivalent transfer in the random pixel setting.}
  \label{fig:svd_alignment}
\end{figure}

Formally, let $W \in \mathbb{R}^{n \times d}$ be the pre-trained weight matrix at a given layer, and consider its singular value decomposition $W = U S V^\top$. The (uncentered) covariance matrix of the pre-trained weights is $C = \frac 1n W^{\top}W = V (\frac1n S^2) V^\top$. Let also $W' = U'S'V'^\top  \in \mathbb{R}^{n \times d}$ represent the initial weights. In the first experiment, we keep only $S$ and $V$ (and thus $C$) and replace $U$ with $U'$. In the complementary experiment, we keep only $U$ and replace both $S$ and $V$ with $S'$ and $V'$ respectively (thereby replacing $C$ with the untrained $C'$).
If the transfer effect persists in the first experiment and disappears in the second experiment, we can attribute all the conv transfer effect to the covariance structure of convolutional weights. An important point is that these SVD replacement operations change the distribution of the hidden activations, leading to a mismatch in the subsequent layer \citep{guth2024rainbow}. To account for and cancel this effect, we apply in both experiments the alignment procedure of \citet{guth2024rainbow}, which restores consistency between layers (details can be found in Appendix \ref{app:alignment_procedure}).
The results are shown in \Cref{fig:svd_alignment}. We find that the transfer disappears when we only keep $U$ (resetting $C$) and that some transfer persists when we keep $S$ and $V$ (thus keeping $C$). We also confirmed that the alignment procedure is important to preserve the transfer when we keep $C$, and it does not provide any advantage if we reset $C$. Noticeable discrepancies between covariance-only and full convolutional transfer however remain---covariance-only transfer is initially more efficient than full convolutional transfer (surprisingly), but saturates with the pre-training time while full transfer continues to decrease $t_0$. This suggests that there might be more precise localizations of transfer than what can be obtained by an SVD, although they remain to be determined. We additionally verified that (i) adding alignment in our previous weight ablation experiments does not change our conclusion  (see \Cref{fig:weight_ablation_random_label_aligned,fig:weight_ablation_random_pixel_aligned}) and (ii) all results in this section are consistent when we reset with freshly drawn weights (see \Cref{fig:weight_ablation_random_label_fresh,fig:weight_ablation_random_pixel_fresh,fig:svd_alignment_fresh_multiseeds}).

\section{Discussion}
\label{discussion}

In this work, we systematically study transfer effects between memorization tasks in a controlled setting. Using $t_0$ as a quantitative measure of transfer, we identify two surprising transfer patterns: equivalent transfer, where one additional pre-training epoch saves one downstream fine-tuning epoch after a warm-up period, and non-equivalent transfer, where pre-training on a mismatched distribution can initially be more efficient than training directly on the downstream task itself. Through weight ablation experiments, we decompose the transfer effects into magnitude-driven transfer for fully-connected weights and structure-driven transfer for convolutional weights, where the latter is partially encoded in their covariance matrix. These effects arise even when the pre-training task itself is completely devoid of structure. 
Our results thus show that the covariance alignment explanation proposed by \citet{maennel2020what} cannot account for all transfer settings, and extend the work of \citet{guth2024rainbow} by demonstrating that the covariance structure encodes beneficial information not only for the upstream task but also for the downstream fine-tuning. They also inform further scientific investigations into transfer learning and suggest simpler pre-training strategies.

There are also some limitations to our work. First, we have limited ourselves to a single architecture (a 7-layer CNN) and base image dataset (CIFAR-10). It would be interesting for future work to explore whether similar phenomena can be observed on other architectures and datasets. Second, our SVD experiments only partially localize the structure-driven transfer (conv-transfer), as the effect of keeping covariance alone saturates with the pre-training time. This indicates that more refined partitions of the learned weights (beyond SVD) may be needed to further localize the structure-driven transfer.

\bibliographystyle{plainnat}
\bibliography{references}

\appendix
\section{Supplementary Results}
\label{sec:supplementary results}

\begin{figure}[t]
  \centering
  \includegraphics[width=\linewidth]{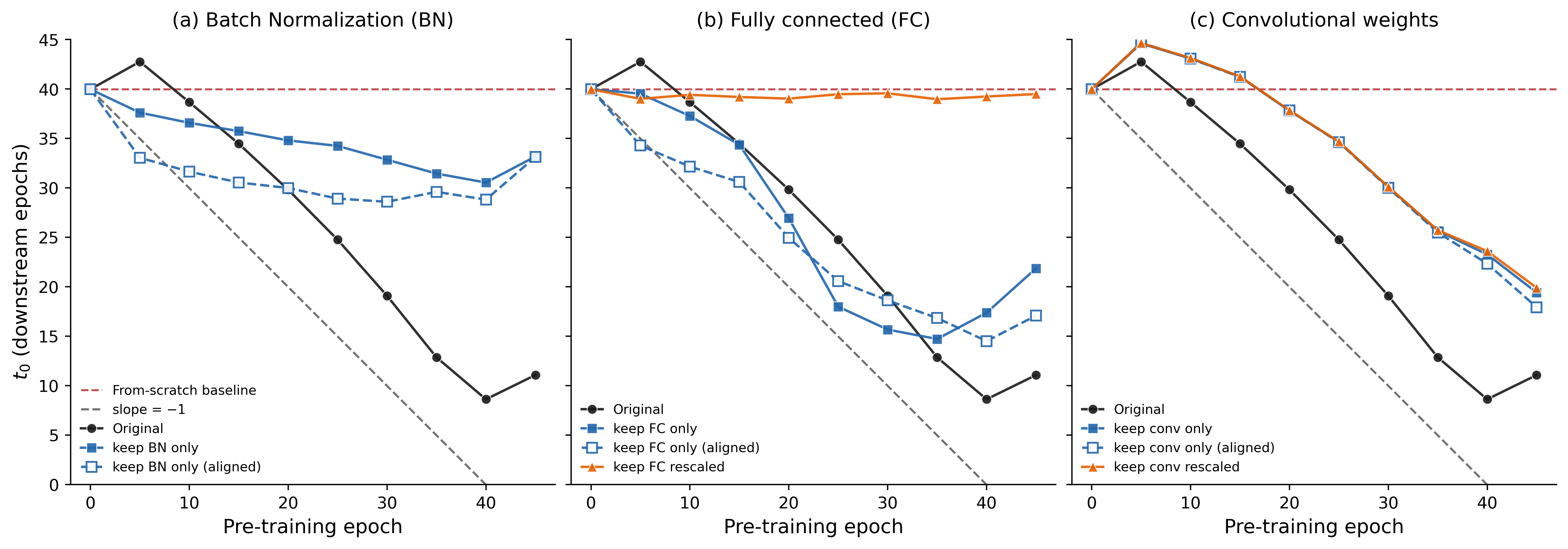}
  \caption{Same experiment as \Cref{fig:weight_ablation_randlabel} with alignment procedure. Across all panels, our conclusion does not change after conducting the alignment procedure.}
  \label{fig:weight_ablation_random_label_aligned}
\end{figure}

\begin{figure}[t]
  \centering
  \includegraphics[width=\linewidth]{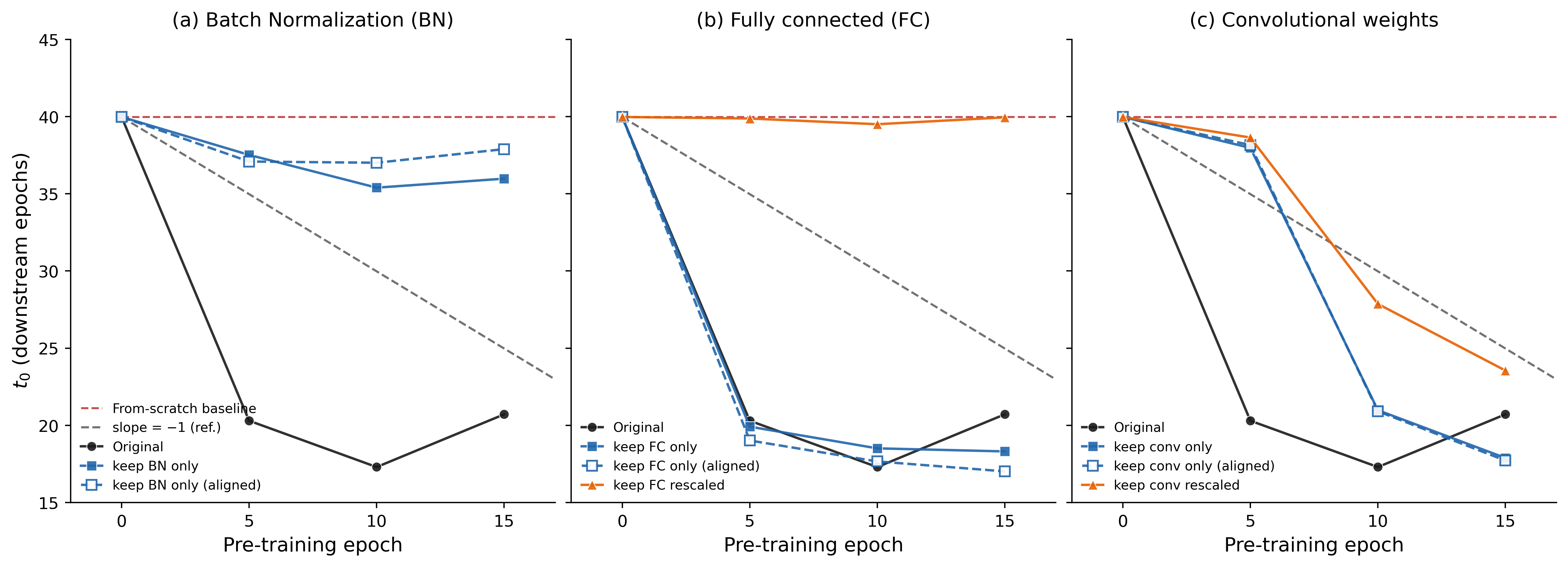}
  \caption{Same experiment as \Cref{fig:weight_ablation_randlabel} with random pixel upstream. \textbf{(a)} keeping $\gamma/\beta$ only preserves negligible transfer effect, with the curve being nearly flat. \textbf{(b)} substantial transfer effect is shown when we only keep the FC weights. After rescaling, the curve becomes nearly flat, suggesting the transfer carried by FC weights can be fully explained by the magnitude of the weights. \textbf{(c)} keeping the convolutional weights alone can also preserve substantial transfer effect. After rescaling, the transfer phenomenon persists, albeit weakened, indicating there is indeed contribution of the growing norm but it is not the only reason for the conv-transfer in random-pixel setting. In all three panels, conducting alignment procedure does not change our conclusion.}
  \label{fig:weight_ablation_random_pixel_aligned}
\end{figure}

\begin{figure}[t]
  \centering
  \includegraphics[width=\linewidth]{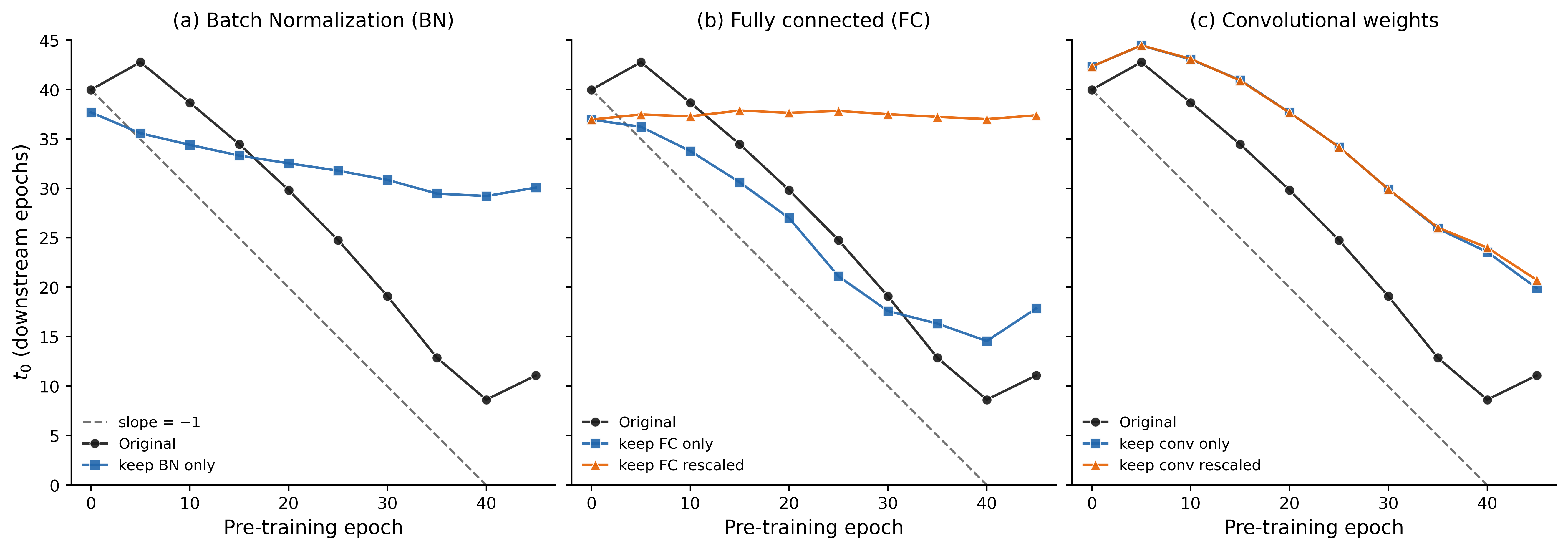}
  \caption{Same experiment as \Cref{fig:weight_ablation_randlabel}, resetting with freshly drawn weights instead of their initial values. Keep-only curves are no longer starting at the same point as the original curve, since now the reset components are drawn from a different seed.}
  \label{fig:weight_ablation_random_label_fresh}
\end{figure}

\begin{figure}[t]
  \centering
  \includegraphics[width=\linewidth]{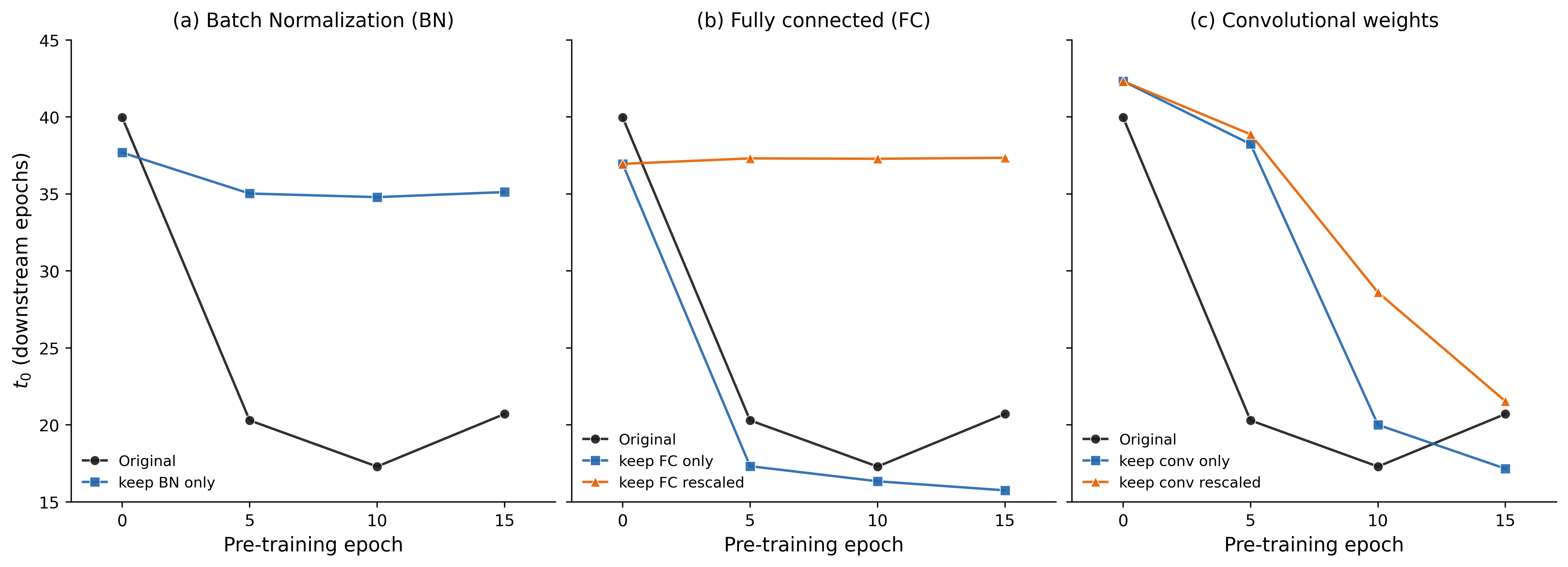}
  \caption{Same experiment as \Cref{fig:weight_ablation_randlabel} with random pixel upstream, resetting with freshly drawn weights.}
  \label{fig:weight_ablation_random_pixel_fresh}
\end{figure}

\begin{figure}[t]
  \centering
  \includegraphics[width=\linewidth]{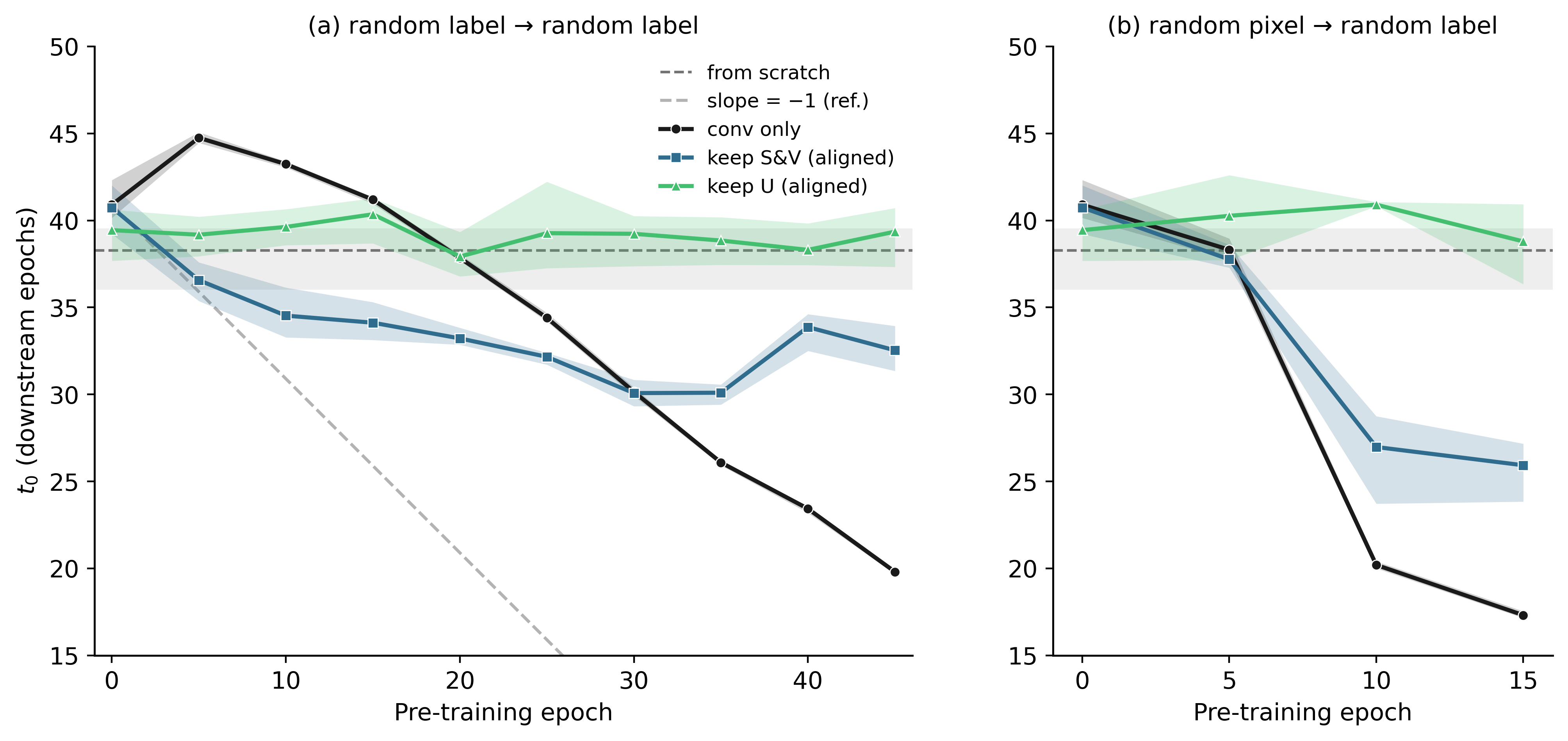}
  \caption{Same experiment as \Cref{fig:svd_alignment}, resetting with freshly drawn weights. The experiment is repeated three times with different seeds. Curves represent the mean of three runs and the shaded area denotes the min--max range.}
  \label{fig:svd_alignment_fresh_multiseeds}
\end{figure}

\clearpage

\section{Experimental Details}
\label{app:exp_details}

\textbf{Architecture:} 6 convolutional layers organized in three blocks with width (128, 256, 512), each layer consisting of Conv2d ($3 \times 3$, stride $=1$, padding $=1$) followed by Batch Norm and ReLU. Each block ends with MaxPool2d ($2 \times 2$). After the final block, global average pooling feeds into a linear classifier ($512 \rightarrow 10$). Convolutional weights are initialized with Kaiming normal; FC weights with $\mathcal{N}(0, 0.01)$.

\textbf{Training:} SGD with a fixed learning rate ($0.01$), momentum $=0.9$, no weight decay, batch size = 128. Training stops at $99.9 \%$. Loss function is cross entropy loss.

\textbf{Data:} CIFAR-10 ($50000$ training images) \citep{krizhevsky2009learning}, normalized with per-channel mean and standard deviation. No data augmentation. Random labels are sampled uniformly from $10$ classes. For random pixels experiments, each pixel is independently sampled from a uniform distribution over $[0, 255]$ and normalized with the same per-channel mean and standard deviation.

\section{Alignment Procedure}
\label{app:alignment_procedure}
In this section, we show how to conduct the alignment procedure of \citet{guth2024rainbow} empirically. Let $\phi_j(x), \phi_j'(x)$ denote the post-BN-ReLU activation of the $j$-th layer for original and modified model respectively.  For each layer, we calculate the cross-covariance matrix $M_j = \sum_x\phi_j(x)\phi_j'(x)^T$ and obtain the alignment matrix $\hat{A}_j = UV^T$ from the SVD of $M_j = USV^T$. This $\hat{A}_j$ is known to minimize $\sum_x \| \hat{A}_j\phi_j'(x) - \phi_j(x) \|^2$ over orthogonal matrices. The alignment is then applied to the next layer's weight as $W_{j+1}' = W_{j+1}\hat{A}_j$, where $W_{j+1}$ is the initial weight of the $(j+1)$-th layer of the modified checkpoint.

\newpage
\section*{NeurIPS Paper Checklist}

\begin{enumerate}

\item {\bf Claims}
    \item[] Question: Do the main claims made in the abstract and introduction accurately reflect the paper's contributions and scope?
    \item[] Answer: \answerYes{} 
    \item[] Justification:  The abstract and introduction state all main claims, including equivalent transfer and non-equivalent transfer phenomenon (Section 2), the localization of transfer and decomposition of the transfer (Section 3), and the random pixel experiment (Section 2).

\item {\bf Limitations}
    \item[] Question: Does the paper discuss the limitations of the work performed by the authors?
    \item[] Answer:  \answerYes{}
    \item[] Justification: The Discussion section discusses limitations including using a single architecture and dataset. And we identified further questions such as if there are higher order statistics that can fully encode the transfer.

\item {\bf Theory assumptions and proofs}
    \item[] Question: For each theoretical result, does the paper provide the full set of assumptions and a complete (and correct) proof?
    \item[] Answer: \answerNA{} 
    \item[] Justification: The paper is empirical and does not include theoretical proofs.

    \item {\bf Experimental result reproducibility}
    \item[] Question: Does the paper fully disclose all the information needed to reproduce the main experimental results of the paper to the extent that it affects the main claims and/or conclusions of the paper (regardless of whether the code and data are provided or not)?
    \item[] Answer: \answerYes{} 
    \item[] Justification: The experiment setup is fully described in the Appendix.

\item {\bf Open access to data and code}
    \item[] Question: Does the paper provide open access to the data and code, with sufficient instructions to faithfully reproduce the main experimental results, as described in supplemental material?
    \item[] Answer: \answerNo{} 
    \item[] Justification: Code will be released with the camera-ready version of the paper (see \Cref{app:exp_details}).

\item {\bf Experimental setting/details}
    \item[] Question: Does the paper specify all the training and test details (e.g., data splits, hyperparameters, how they were chosen, type of optimizer) necessary to understand the results?
    \item[] Answer: \answerYes{} 
    \item[] Justification: All details are included in Appendix.

\item {\bf Experiment statistical significance}
    \item[] Question: Does the paper report error bars suitably and correctly defined or other appropriate information about the statistical significance of the experiments?
    \item[] Answer: \answerYes{} 
    \item[] Justification: Key experiments are repeated across multiple seeds. The equivalent transfer relationship is validated across four seeds (\Cref{fig:t0_linear}), while the SVD replacement experiment (\Cref{fig:svd_alignment}) is repeated across three seeds. Supporting experiments, such as norm rescaling, are intended as demonstrations and are therefore reported using representative seeds.


\item {\bf Experiments compute resources}
    \item[] Question: For each experiment, does the paper provide sufficient information on the computer resources (type of compute workers, memory, time of execution) needed to reproduce the experiments?
    \item[] Answer: \answerNo{} 
    \item[] Justification: We did not record exact compute times. All experiments use a 7-layer CNN on CIFAR-10 using NVIDIA RTX 4090 and RTX 5090 GPUs.

    
\item {\bf Code of ethics}
    \item[] Question: Does the research conducted in the paper conform, in every respect, with the NeurIPS Code of Ethics \url{https://neurips.cc/public/EthicsGuidelines}?
    \item[] Answer: \answerYes{} 
    \item[] Justification: This work complies with the NeurIPS Code of Ethics.


\item {\bf Broader impacts}
    \item[] Question: Does the paper discuss both potential positive societal impacts and negative societal impacts of the work performed?
    \item[] Answer: \answerNA{} 
    \item[] Justification: This is foundational research aimed at understanding transfer in neural networks. We do not anticipate any direct negative societal impacts.

    
\item {\bf Safeguards}
    \item[] Question: Does the paper describe safeguards that have been put in place for responsible release of data or models that have a high risk for misuse (e.g., pre-trained language models, image generators, or scraped datasets)?
    \item[] Answer: \answerNA{} 
    \item[] Justification: This paper does not release any model or dataset that can be misused.

\item {\bf Licenses for existing assets}
    \item[] Question: Are the creators or original owners of assets (e.g., code, data, models), used in the paper, properly credited and are the license and terms of use explicitly mentioned and properly respected?
    \item[] Answer: \answerYes{} 
    \item[] Justification: We use CIFAR-10, which is publicly available for research and properly cited in our paper.


\item {\bf New assets}
    \item[] Question: Are new assets introduced in the paper well documented and is the documentation provided alongside the assets?
    \item[] Answer: \answerNA{} 
    \item[] Justification: The paper does not release new datasets or models.

\item {\bf Crowdsourcing and research with human subjects}
    \item[] Question: For crowdsourcing experiments and research with human subjects, does the paper include the full text of instructions given to participants and screenshots, if applicable, as well as details about compensation (if any)? 
    \item[] Answer: \answerNA{} 
    \item[] Justification: This paper does not involve human subjects or crowdsourcing.


\item {\bf Institutional review board (IRB) approvals or equivalent for research with human subjects}
    \item[] Question: Does the paper describe potential risks incurred by study participants, whether such risks were disclosed to the subjects, and whether Institutional Review Board (IRB) approvals (or an equivalent approval/review based on the requirements of your country or institution) were obtained?
    \item[] Answer: \answerNA{} 
    \item[] Justification: This study does not involve human subjects research.


\item {\bf Declaration of LLM usage}
    \item[] Question: Does the paper describe the usage of LLMs if it is an important, original, or non-standard component of the core methods in this research? Note that if the LLM is used only for writing, editing, or formatting purposes and does \emph{not} impact the core methodology, scientific rigor, or originality of the research, declaration is not required.
    \item[] Answer: \answerNA{} 
    \item[] Justification: LLM is not used as a component of the key methodology.

\end{enumerate}

\end{document}